\documentclass[sigconf]{acmart}
\AtBeginDocument{%
  \providecommand\BibTeX{{%
    \normalfont B\kern-0.5em{\scshape i\kern-0.25em b}\kern-0.8em\TeX}}}

\setcopyright{acmcopyright}

\copyrightyear{2021}
\acmYear{2021}
\setcopyright{acmcopyright}\acmConference[IJCKG'21]{The 10th International Joint Conference on Knowledge Graphs}{December 6--8, 2021}{Virtual Event, Thailand}
\acmBooktitle{The 10th International Joint Conference on Knowledge Graphs (IJCKG'21), December 6--8, 2021, Virtual Event, Thailand}
\acmPrice{15.00}
\acmDOI{10.1145/3502223.3502244}
\acmISBN{978-1-4503-9565-6/21/12}

\usepackage{microtype}

\usepackage{amsmath}
  
\usepackage{amssymb}
\usepackage{multirow}
\usepackage{multicol}
\usepackage{booktabs}
\usepackage{arydshln}
\usepackage{graphicx}
\usepackage{comment}

\begin{document}


\title{Improving Knowledge Graph Representation Learning by Structure Contextual Pre-training}


\author{Ganqiang Ye}
\affiliation{%
  \institution{Zhejiang University}
  \city{Hangzhou, Zhejiang}
  \country{China}
  }
\email{yeganqiang@zju.edu.cn}

\author{Wen Zhang}
\affiliation{%
  \institution{Zhejiang University}
  \city{Hangzhou, Zhejiang}
  \country{China}
  }
\email{wenzhang2015@zju.edu.cn}

\author{Zhen Bi}
\affiliation{%
  \institution{Zhejiang University}
  \city{Hangzhou, Zhejiang}
  \country{China}
  }
\email{bizhen_zju@zju.edu.cn}

\author{Chi Man Wong}
\affiliation{%
  \institution{Alibaba Group}
  \city{Hangzhou, Zhejiang}
  \country{China}
  }
\email{chiman.wcm@alibaba-inc.com}

\author{Hui Chen}
\affiliation{%
  \institution{Alibaba Group}
  \city{Hangzhou, Zhejiang}
  \country{China}
  }
\email{weidu.ch@alibaba-inc.com}

\author{Huajun Chen}
\authornote{Corresponding author}
\affiliation{%
  \institution{Zhejiang University \& AZFT JointLab for Knowledge Engine}
  \city{Hangzhou, Zhejiang}
  \country{China}
  }
\email{huajunsir@zju.edu.cn}

\renewcommand{\shortauthors}{Ganqiang Ye and Zhang Wen, et al.}

\begin{abstract}
Representation learning models for Knowledge Graphs (KG) have proven to be effective in encoding structural information and performing reasoning over KGs. In this paper, we propose a novel \emph{pre-training-then-fine-tuning} framework for knowledge graph representation learning, in which a KG model is firstly pre-trained with triple classification task, followed by discriminative fine-tuning on specific downstream tasks such as entity type prediction and entity alignment. Drawing on the general ideas of learning deep contextualized word representations in typical pre-trained language models, we propose SCoP to learn pre-trained KG representations with structural and contextual triples of the target triple encoded. 
Experimental results demonstrate that fine-tuning SCoP not only outperforms results of baselines on a portfolio of downstream tasks but also avoids tedious task-specific model design and parameter training.
\end{abstract}



\maketitle

\section{Introduction}
Knowledge Graphs (KGs) can be regarded as directed labeled graphs in which facts are represented as triples in the form of \textit{(head entity, relation, tail entity)}, abbreviated as \textit{(h,r,t)}. In recent years, KGs gain rapid development on both construction and applications, proven to be useful in 
artificial intelligence tasks such as semantic search \citep{Paraphrasing-Berant-2014}, information extraction \citep{Multilingual-Daiber-2013}, and question answering \citep{QA-Diefenbach-2018}.

Graph structure in KGs contains a lot of valuable information \citep{GraRep-Cao-2015, RLG-Hamilton-2017}, 
thus representation learning methods, embedding 
entities and relations into continuous vector spaces, are proposed to extract these structural features \citep{TransE-Bordes-2013, TransH-Wang-2014, TransR-Lin-2015}.

Various representation learning methods are specially proposed for different tasks, such as entity type prediction \citep{FIGMENT-Yaghoobzadeh-2015}, entity alignment \citep{ITransE-Zhu-2017} and link prediction \citep{SimplE-Kazemi-2018}. 
In our opinion,
though tasks are different, KGs might be the same. 
Thus a general way to encode KGs is necessary. 
And the challenge is to 
make structural and contextual information into consideration, since they varies in different tasks. 

Inspired by Pre-trained Language Models (PLM) such as BERT \citep{BERT-Devlin-2019}, which learn deep contextual representations for words and has achieved significant improvement in a variety of NLP tasks, we propose a novel knowledge graph representation learning method by Structure Contextual Pre-training (SCoP) model, to extract deep structural features and context 
for different knowledge graph tasks automatically. Similar to 
PLMs, there are two steps for using SCoP, pre-training and fine-tuning. Pre-training step enables the model to extract various deep structural contextualized information of entities and relations in knowledge graph and fine-tuning step makes the model adapt to specific downstream tasks.

With a target triple $(h,r,t)$ as input, SCoP 
mainly includes two parts, 
in which contextual module generates representations for $h, r, $ and $t$'s  contextual triples,
and aggregation module uses Transformer \cite{Attention-Vaswani-2017} to learn structural features with all contextual triples 
and fully interacted and generates a task-specific output vector. 

We construct a new KG dataset, WN-all, which contains all the triple from WordNet and is suitable for KG pre-training model. We experimentally prove that SCoP is capable of doing different downstream tasks including entity type prediction, entity alignment and triple classification, and it could achieve better results in an efficient way. What's more, SCoP exhibits good prediction and classification performance and the importance of structural and contextual information in KG.

\section{SCoP}

\subsection{SCoP overview}
Similar to pre-trained language model, SCoP model in this paper includes two steps: pre-training and fine-tuning. The SCoP model is based on structure contextual triples to learn the structural information of knowledge graph.

From the perspective of model structure, SCoP consists two modules that integrate and extract the deep structure information of KG. In this paper, bold characters denote vector representations, for example, $\mathbf{h}$ is the vector representation of entity $h$.

\begin{figure}[tp]
  \centering
  \includegraphics[width=0.5\textwidth]{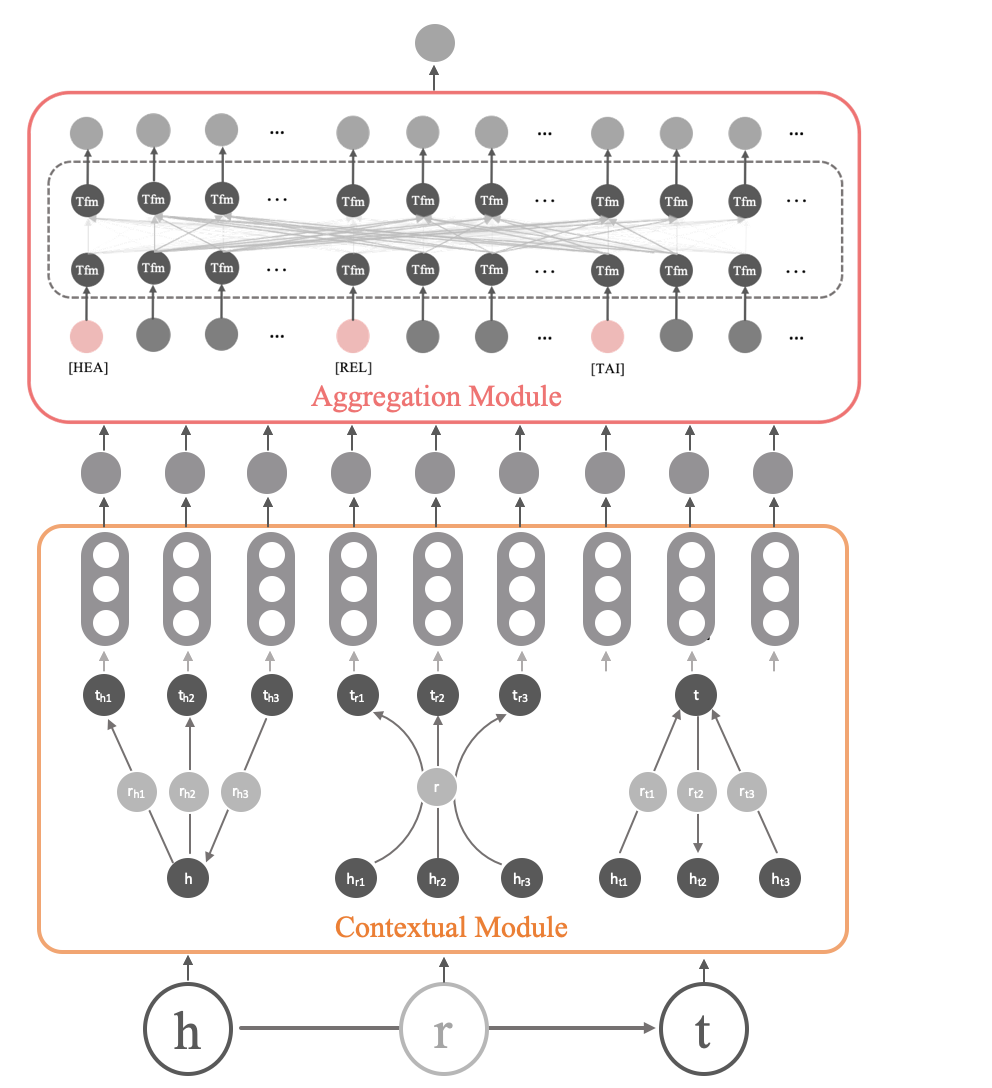}
  \caption{The structure of SCoP model. Given a target triple $(h, r, t)$, find its contextual triples and input them into SCoP model through two modules. Finally, we get the aggregation output representation.}
  \label{structure-scop}
\end{figure}

\begin{table*}[tp]
\caption{\label{tab:type-prediction-and-alignment-performance}
The results of entity type prediction task and entity alignment task. The bold number is the best one among all results while underlined number is the second best. Suffixes "-pr" and "-ft" means the pre-training and fine-tuning process respectively.
}
  \centering
  {\tabcolsep0.13in
  \small
    \begin{tabular}{lcccccccc}
    \toprule
    \multirow{2}{*}{\textbf{Method}}&
    \multicolumn{4}{c}{\textbf{Entity Type Prediction}}&\multicolumn{4}{c}{\textbf{Entity Alignment}}\cr
    \cmidrule(lr){2-5} \cmidrule(lr){6-9}
    & MRR & Hit@1 & Hit@3 & Hit@10 & MRR & Hit@1 & Hit@3 & Hit@10\cr
    \midrule
    TransE-pr
    	& 0.8499 & 83.11 & 86.35 & 87.75 & 0.8964 & 88.65 & 90.08 & 91.06  \cr
    TransE-ft
    	& 0.8386 & 81.99 & 85.10 & 86.92 & 0.8943 & 88.63 & 89.76 & 90.89   \cr
	ComplEx-pr
		& 0.8396 & 80.30 & 86.77 & 89.15 & 0.8636 & 82.27 & 90.03 & 90.94 \cr
    ComplEx-ft
    	& 0.7809 & 72.09 & 77.49 & 82.72 & 0.8772 & 84.26 & 89.56 & 90.23  \cr
    RotatE-pr
    	& 0.8941 & 88.12 & \underline{90.93} & \underline{91.62} & 0.9066 & 89.78 & 91.34 & 92.20 \cr
    RotatE-ft
    	& 0.8814 & 86.60 & 90.05 & 90.88 & 0.9035 & 89.75 & 90.77 & 91.36  \cr
    \midrule
    SCoP-pr & \underline{0.9018} & \underline{89.73} & 90.11 & 90.37 & \underline{0.9271} & \underline{92.61} & \underline{92.78} & \underline{92.88} \cr
    SCoP-ft & \bf{0.9208} & \bf{91.98} & \bf{92.06} & \bf{92.51} & \bf{0.9319} & \bf{93.15} & \bf{93.21} & \bf{93.39} \cr
    \bottomrule
   \end{tabular}
   }

\end{table*}

\subsection{Structure Contextual Triples}
Given a KG $\mathcal{G} = \{ \mathcal{E}, \mathcal{R}, \mathcal{T}\}$, $\mathcal{E}$, $\mathcal{R}$ and $\mathcal{T}$ are the set of entities, relations and triples respectively.
And $\mathcal{T} = \{(h,r,t) |h\in \mathcal{E}, r\in \mathcal{R}, t \in \mathcal{E} \}$.

\paragraph{Entity contextual triples}
For an entity $e$, contextual triples of it $C(e) \subseteq \mathcal{T}$  are triples containing $e$ either as head or tail entity. Specifically, given an entity $e$, the contextual triple set of $e$ is
\begin{align*}
C(e) =\hspace*{0.1cm}&\{ {(e,r,t\hspace*{0.07cm})|(e,r,t\hspace*{0.07cm}) \in \mathcal{T},\hspace*{0.08cm}t \in \mathcal{E}, r \in \mathcal{R}} \}\hspace*{0.1cm}\cup \\
&\{ {(h,r,e)|(h,r,e) \in \mathcal{T}, h \in \mathcal{E},r \in \mathcal{R}} \}
\end{align*}

\paragraph{Relation contextual triples}
Similarly, triples that contain the same relation $r$ are the contextual triples of $r$. Unlike entity contextual triples, relation contextual triples are not directly adjacent. Given a relation $r$, $C(r)$ is
\[C(r) = \left\{ {(e_1,r,e_2)|(e_1,r,e_2) \in \mathcal{T}, \ e_1,e_2 \in \mathcal{E}} \right\}\]

\subsection{Two Steps: Pre-training and Fine-tuning}
At the pre-training step, all parameters of the model are first trained on a general task which is triple classification in this paper, as the existence of a triple is the most widely available information in a knowledge graph.

\begin{figure}[tp]
  \centering
  \includegraphics[width=0.5\textwidth]{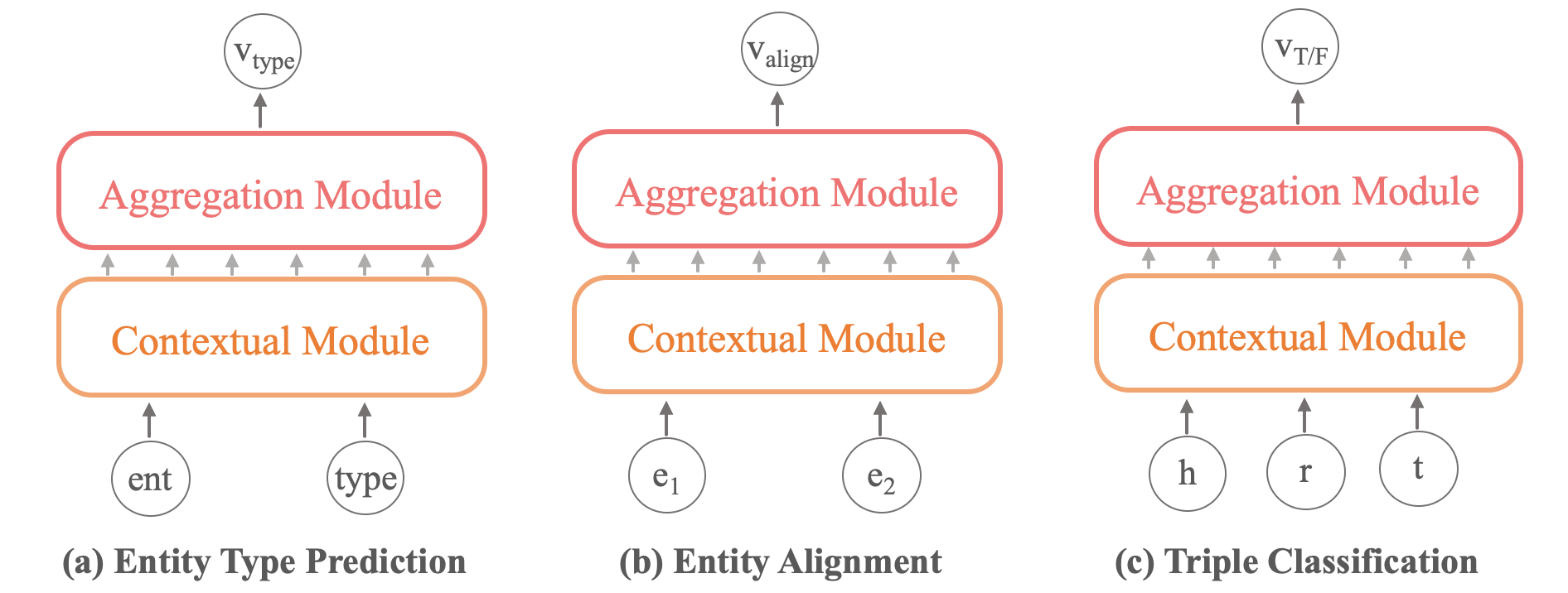}
  \caption{The structure of three model variants at the fine-tuning step, correspond to three downstream tasks.}
  \label{fine-tuning-structure}
\end{figure}

At the fine-tuning step, we generate model variants based on the specific downstream task and the parameters in SCoP are fine-tuned with its dataset, as shown in Figure \ref{fine-tuning-structure}. This process is lightweight and fast to better results. 

\subsection{Two Modules}

\paragraph{Contextual module}
Given a target triple $\tau = (h,r,t)$, Contextual Module (C-Mod) first finds contextual triples of $h$, $r$ and $t$, and then encodes each contextual triple $ c = (h^\prime, r^\prime, t^\prime) \in \{C(h), C(r), C(t) \}$ into a triple representation $\mathbf{c}$. Thus
$$ \mathbf{c} = \text{C-Mod}\hspace*{0.06cm}(<\mathbf{h}^\prime, \mathbf{r}^\prime, \mathbf{t}^\prime>) $$
where $<\mathbf{x},\mathbf{y},\mathbf{z}>$ denotes a sequence of $\mathbf{x},\mathbf{y}$ and $\mathbf{z}$.
We implement the triple module based on a single layer feed-forward neural network.

\paragraph{Aggregation module}

Aggregation Module (A-Mod) takes contextual triple representation sequences and outputs an aggregation representation $\mathbf{a}$, so A-Mod can be expressed as
$$ \mathbf{a} = \text{A-Mod}\hspace*{0.06cm}(seq_{h}, seq_{r}, seq_{t}) $$
where $seq_{h}, seq_{r}, seq_{t}$ are the sequence of contextual triples of $h, r, t$ represented as follows
$$seq_x = < c_x^1,  c_x^2, ...,  c_x^n> $$
and $c_x^i$ is the $i$-th contextual triples of $x$ where $x$ is one of $h, r, t$. $n$ is the number of contextual triples.

We introduce segment vector $\mathbf{s}_{x}$ into the current contextual triple sequences as Bert \citep{BERT-Devlin-2019}, indicating whether it belongs to head entity $h$, relation $r$ or tail entity $t$.
Besides, three types of triple token \texttt{[HEA]}, \texttt{[REL]} and \texttt{[TAI]} are added in front of each contextual triple of $h,r,t$ and their representations are denoted as $\mathbf{k}_{\text{x}}$.
Consequently, the updated input sequence can be represented as
$$ \mathbf{i} = <\mathbf{k}_{\text{[HEA]}}, \widehat{\mathbf{seq}}_{h}, \mathbf{k}_{\text{[REL]}}, \widehat{\mathbf{seq}}_{r}, \mathbf{k}_{\text{[TAI]}}, \widehat{\mathbf{seq}}_{t}> $$
where $\widehat{\mathbf{seq}}_{x}$ is the sequence of updated contextual triple representations $\mathbf{c}_x^i$.
A-Mod encodes the input sequence $\mathbf{i}$ with multi-layer bidirectional Transformer encoder, referred to the key idea of Bert.


Eventually, based on the three structure representations, $\mathbf{h}_{s}$, $\mathbf{r}_{s}$ and $\mathbf{t}_{s}$, the scoring function 
can be written as 


  $$s_{\tau} = f(h,r,t) = {\rm softmax}(([\mathbf{h}_{s};\mathbf{r}_{s};\mathbf{t}_{s}] \hspace*{0.1cm} \mathbf{W}_{int} + \mathbf{b}) \mathbf{W}_{cls})$$
where $[\mathbf{x};\mathbf{y};\mathbf{z}]$ denotes the concatenation vector $\mathbf{x}$, $\mathbf{y}$ and $\mathbf{z}$. $\mathbf{W}_{int} \in \mathbb{R}^{3d \times d}$ and $\mathbf{W}_{cls} \in \mathbb{R}^{d \times 2}$ are weights while $\textbf{b} \in \mathbb{R}^d$ is a bias vector. 
$s_{\tau} \in \mathbb{R}^2$ is a 2-dimensional real vector with
$s_{\tau_0} + s_{\tau_1} = 1$ after softmax function.
Given the positive triple set $\mathbb{D}^+$ and a negative triple set $\mathbb{D}^-$ constructed accordingly, we compute a cross-entropy loss with $s_{\tau}$ and triple labels:
$$ \mathcal L= \sum\limits_{\tau  \in \mathbb{D}^+ \cup \mathbb{D}^-} {(\hspace*{0.1cm}{y_\tau }\hspace*{0.1cm}{\rm{log}}({s_{{\tau _0}}}) + (1 - {y_\tau })\hspace*{0.05cm}{\rm{log}}({s_{{\tau _1}}})\hspace*{0.05cm})} $$
where $y_\tau \in \{0, 1\}$ is the label of $\tau$, and is $1$ if $\tau \in \mathbb{D}^+$ while $0$ if $\tau \in \mathbb{D}^-$. The negative triple set $\mathbb{D}^-$ is simply generated by
replacing head entity $h$ or tail entity $t$ with another random entity $e\in \mathcal{E}$ or replacing relation $r$ with another random relation $r^\prime \in \mathcal{R}$.

\begin{table}\footnotesize%
\caption{\label{tab:datasets} Statistics of datasets.}
\centering
\resizebox{.5\textwidth}{!}{
\begin{tabular}{lccccc}
\toprule
\textbf{Dataset}  & \textbf{\# Ent}  & \textbf{\# Rel} & \textbf{\# Train} & \textbf{\# Dev} & \textbf{\# Test} \\
\midrule
WN-all  & 116,377 & 26 & 302,074 & 37,759 & 37,759 \\
\midrule
Type-Prediction    & 87,875 & - &  72,076 & 8,431 & 8,582 \\
Entity-Alignment & 13,205 & - &  17,725 & 1,805 & 1,856 \\
\bottomrule
\end{tabular}
}
\end{table}


\section{Experiments}

\subsection{Datasets}
To prove the effective of pre-training knowledge graph, we conduct experiments
on the whole knowledge graph WordNet, named WN-all. 
We also construct two downstream datasets for fine-tuning. 
Detailed statistics of datasets are shown in Table \ref{tab:datasets}. 

\subsection{Pre-training SCoP}\label{pre-trainin-scop}
We pre-train SCoP on WN-all via triple classification task,
aiming to judge whether the triple is correct with label $l$. The loss is defined as 
$ \mathcal L= CrossEntropy(log(softmax(\mathbf{a})), l)$.

We fix the length of input contextual triple sequence with 256 and the length of each contextual triple of $h$, $r$, $t$ is 84.
We set the self-attention with 6 layers, 3 heads and the hidden dimension of representation vectors of 192 in Transformer in A-Mod, and learning rate of $2e^{-5}$, training batch size of 32, learning rate warm-up over the first 1,000 steps and a dropout probability of 0.1 on all layers.

\begin{figure*}[t]
  \centering
  \includegraphics[width=1.0\textwidth]{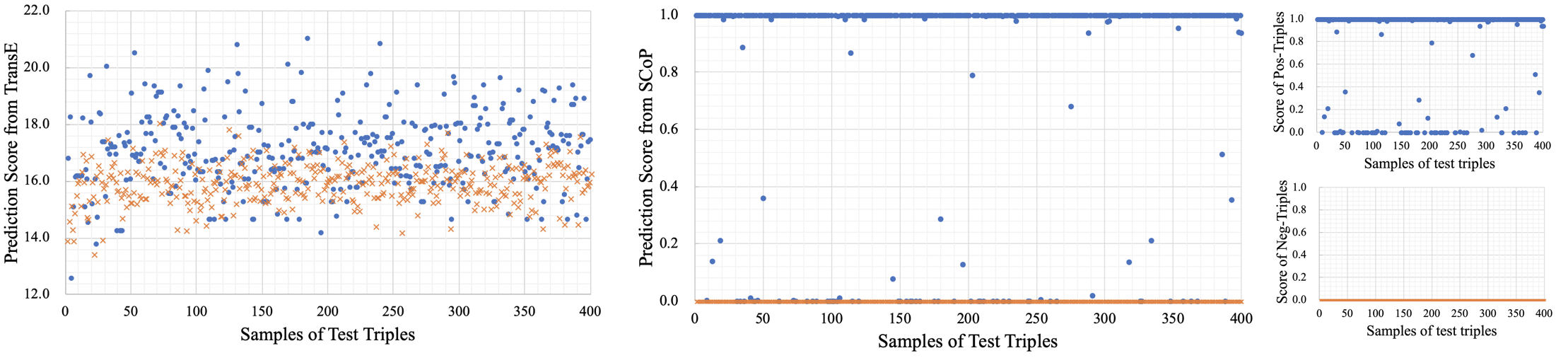}
  \caption{
  The prediction score distribution of TransE and SCoP. The blue dots represent positive sample predictions, while the orange crosses represent negative samples. To show the distribution of SCoP more clearly, the score of positive and negative samples is shown in the two small graphs juxtaposed on the right.
  }
  \label{prob_distribution}
\end{figure*}

\subsection{Entity type prediction}\label{entity-type-prediction}

Entity type prediction concentrates to complete a pair (entity, entity type) when its type is missing, which aims to verify the capability of our model for inferring missing entity type instances. Under this definition, we extract entity and its type from triples with relation $ \_hypernym$ in WN-all as the dataset for this task, shown in Table \ref{tab:datasets}.

To fine-tune on entity type prediction task and its dataset, we input the structure contextual triples sequence of entity type pair $\tau$ and use the output vector $\mathbf{a_{\tau}} \in \mathbb{R}^{d}$ of A-Mod to compute the score function 
$s_{\tau} = f(\rm{ent}, \rm{type}) = {\rm softmax}(\mathbf{a}_{\tau} \mathbf{W})$
and entity type classification loss: 
$ \mathcal L_{\tau}= CrossEntropy(log(s_{\tau}), l_{\tau})$,
where $\mathbf{W} \in \mathbb{R}^{d \times 2}$ is the classification weight and $s_{\tau} \in \mathbb{R}^2$ is a 2-dimensional real vector. Label $l_\tau \in \{0, 1\}$ is $1$ if entity type pair $\tau$ is positive while $0$ if negative. The Figure \ref{fine-tuning-structure} (a) shows the structure of fine-tuning SCoP for entity type prediction task.

We adopt the Mean Reciprocal Rank(MRR) and percentage of triples ranked in top $k$ (Hit@$k$) where $k \in \{1,3,10\}$ as evaluation metrics, which are commonly adopted in previous KG works. See Section \ref{Knowledge-Graph-Embedding} for details of baseline models.

Experimental results are presented in the left part of Table \ref{tab:type-prediction-and-alignment-performance}. Compared with the results of fine-tuning stage (even rows with "-ft" suffix in table), SCoP-ft outperforms other models on all evaluation metrics, especially obtaining 3.8\% prediction accuracy improvement on Hit@1 over the prior state of the art. Besides, we observe that SCoP-ft model has better performance than SCoP-pr while on the contrary, the results of other models after fine-tuning decreased. It proves that the structure contextual triples, and the structural information that they represent in essence, can effectively improve the effect of entity type prediction.

\subsection{Entity alignment}\label{entity-alignment}

Entity alignment task aims to find two entities that refer to the same thing in the real world. Similarly, we generate the dataset for entity alignment consisting of two entities with the same meaning from triples with relation $\_similar\_to$, such as (nascent, emergent) and (sleep, slumber).
Similar to the fine-tuning progress of entity type prediction, we compute the score function $s_{\omega}$ for alignment entity pair $\omega$ and its classification loss $ \mathcal L_{\omega}$ in the same way. Figure \ref{fine-tuning-structure} (b) shows its model structure and the right part of Table \ref{tab:type-prediction-and-alignment-performance} shows the results.

The experimental results show that SCoP model still has the greatest advantage in Hit@1 metric, which is 93.15\% and surpasses the second result by 3.3\%. On the entity type prediction task, the results from pre-training step of SCoP model are not always optimal. For example, RotatE-pr has a better result with Hit@10 metric on entity type prediction task. However, those results of SCoP model outperform all other models at pre-training step on entity alignment task, let alone the overwhelming results after fine-tuning.

\subsection{Triple classification and prob-distribution}\label{triple-classification}

We observed that the data of Hit@1, Hit@3, and Hit@10 in SCoP model are very close, and the interval between them is less than 1\% while those of other models have clear gaps. Therefore, we further investigated the distribution of prediction score and the triple classification task, whose aim is to judge whether a given triple ($h,r,t$) is correct or not, as shown in Figure \ref{fine-tuning-structure} (c).

From Figure \ref{prob_distribution}, it is not difficult to see that the prediction scores of the positive samples and negative samples are relatively evenly distributed in the whole score space predicted by the baseline model like TransE, and most of the positive examples are in the upper part of the space while the negative in the lower part. The score distribution of ComplE and RotatE resembles TransE, so these similar figures are not repeatedly shown. However, The prediction scores of positive and negative samples in SCoP model are close to their upper and lower boundaries respectively and only a small part of the predicted values are distributed in the middle region. Figure \ref{prob_distribution} shows that structural information can bring significant gains to triple classification tasks, and it is easier to set the margin $\gamma$ to distinguish between positive and negative samples. The experiment shows that, different margins $\gamma$ among $\{20, 40, 60, 80\}$ percent of maximum prediction score, hardly affect the prediction accuracy of SCOP with 92.6\% numerically, while that of TransE varies greatly, from 13.6\% to 91.1\%.

\section{Related Work}

\paragraph{Knowledge Graph Embedding (KGE) }\label{Knowledge-Graph-Embedding}
With the introduction of the embedding vector into Knowledge Graph by TransE model \citep{TransE-Bordes-2013}, the method of KGE has made great progress. The key idea is to transform entities and relations in triple into continuous vector space.
ComplEx \citep{trouillon2016complex} introduces complex embeddings so as to better model asymmetric relations and RotatE \citep{sun2019rotate} further infers the composition pattern.

\paragraph{Pre-train Language Models (PLM)}
The PLMs have achieved excellent results in many natural language processing (NLP) tasks, such as BERT \citep{BERT-Devlin-2019}. Later studies have proposed variants of BERT combined with KG, like K-BERT \citep{K_BERT-Liu-2019}, KG-BERT \citep{KG_BERT-Yao-2019}, ERNIE \citep{ERNIE-Zhang-2019}
and KnowBert \citep{KnowBert-Peters-2019}, paying attention to the combination of knowledge graph and pre-trained language model.
However, these methods aim at adjusting the PLM by injecting knowledge bases or their text representations. Whereas, our SCoP model mainly focuses on encoding knowledge graph instead of texts in sentences, so as to extract deep structural information to apply to various downstream tasks in knowledge graph.

\section{Conclusion}
In this work, we construct a new dataset WN-all from complete triples in WordNet and introduce SCoP, a pre-training knowledge graph representation learning method that achieves state-of-the-art results on three various knowledge graph downstream tasks on this dataset. SCoP shows good prediction and classification performance, due to its ability to capture the structural information in KG dynamically. In the future, we plan to evaluate SCoP in more scenarios and to fine-tune it in more different ways.


\begin{acks}
This work is funded by national key
research program 2018YFB1402800, and NSFC91846204/U19B2027.
\end{acks}

\bibliographystyle{ACM-Reference-Format}
\bibliography{sample-base}

\end{document}